\documentclass[conference]{IEEEtran}
\IEEEoverridecommandlockouts
\usepackage{cite}
\usepackage{amsmath,amssymb,amsfonts}
\usepackage{algorithmic}
\usepackage{graphicx}
\usepackage{textcomp}
\usepackage{xcolor}

\usepackage{multirow}
\usepackage{booktabs}
\usepackage[ruled,linesnumbered]{algorithm2e}
\usepackage[normalem]{ulem}
\usepackage{balance}
\usepackage{threeparttable}
\usepackage{url}

\def\BibTeX{{\rm B\kern-.05em{\sc i\kern-.025em b}\kern-.08em
    T\kern-.1667em\lower.7ex\hbox{E}\kern-.125emX}}
\begin{document}

\title{DeepTPI: Test Point Insertion with Deep Reinforcement Learning}

\author{\IEEEauthorblockN{Zhengyuan Shi\IEEEauthorrefmark{1}\textsuperscript{\textsection}, Min Li\IEEEauthorrefmark{1}\textsuperscript{\textsection}, Sadaf Khan\IEEEauthorrefmark{1}, Liuzheng Wang\IEEEauthorrefmark{2}, Naixing Wang\IEEEauthorrefmark{2}, Yu Huang\IEEEauthorrefmark{2} and Qiang Xu\IEEEauthorrefmark{1}}
\IEEEauthorblockA{\IEEEauthorrefmark{1}\textit{Department of Computer Science and Engineering},
\textit{The Chinese University of Hong Kong},
Shatin, Hong Kong S.A.R.\\
\{zyshi21, mli, skhan, qxu\}@cse.cuhk.edu.hk}

\IEEEauthorblockA{\IEEEauthorrefmark{2}\textit{HiSilicon Technologies Co., Ltd.}, China \\
}
}

\maketitle

\begingroup\renewcommand\thefootnote{\textsection}
\footnotetext{Both authors contributed equally to this research.}
\endgroup

\begin{abstract}
Test point insertion (TPI) is a widely used technique for testability enhancement, especially for logic built-in self-test (LBIST) due to its relatively low fault coverage. 
In this paper, we propose a novel TPI approach based on deep reinforcement learning (DRL), named \emph{DeepTPI}. Unlike previous learning-based solutions that formulate the TPI task as a supervised-learning problem, we train a novel DRL agent, instantiated as the combination of a graph neural network (GNN) and a Deep Q-Learning network (DQN), to maximize the test coverage improvement. Specifically, we model circuits as directed graphs and design a graph-based value network to estimate the action values for inserting different test points. The policy of the DRL agent is defined as selecting the action with the maximum value.
Moreover, we apply the general node embeddings from a pre-trained model to enhance node features, and propose a dedicated testability-aware attention mechanism for the value network. 
Experimental results on circuits with various scales show that DeepTPI significantly improves test coverage compared to the commercial DFT tool. The code of this work is available at \url{https://github.com/cure-lab/DeepTPI}.
\end{abstract}

\section{Introduction} \label{Sec:Intro}

Logic built-in-self-test (LBIST)~\cite{mccluskey1985built} is a commonly used technique for in-field tests, especially for many safety-critical scenarios~\cite{he2017test}, such as self-driving cars, healthcare systems, and aircraft operating systems that have strict reliability requirements. 
LBIST uses a pseudo-random pattern generator to simulate the circuit behavior and verifies the corresponding responses. While LBIST demands low hardware and computational overhead, it fails to detect the random pattern resistant (RPR) faults~\cite{mccluskey1985built}. 
The RPR fault commonly occurs in complex modern circuits, where the gate has many fan-in wires. In most cases, there is only one valid input assignment out of the quadratic fan-in number of total random combinations. Therefore, the existence of the RPR faults results in the insufficient test coverage under pseudo-random test.

To improve the low test coverage of LBIST caused by RPR faults, test point insertion (TPI)~\cite{williams1973enhancing} has been studied extensively, which inserts extra gates, i.e., test points (TPs), into the circuit during the design-for-test (DFT) phase. As finding an optimal TP position is a known \textit{NP-hard} problem~\cite{krishnamurthy1987dynamic}, many methods have been proposed to build computationally feasible TPI algorithms. 
The straight-forward approaches perform fault simulation to identify the RPR faults and then insert TPs targeting on such faults~\cite{williams1973enhancing, tamarapalli1996constructive, iyengar1989synthesis}. 
The approximate-measurement approaches estimate the testability by fast approximation and insert TPs to improve testability~\cite{COP1984OnTO, touba1996test, das2000reducing}. However, both of them are often ineffective on modern circuits. The fault simulation approach requires the unacceptable runtime proportional to the size of the circuit~\cite{tsai1997hybrid}, while the accuracy of approximate measurement usually degrades due to the reconvergence structures~\cite{roy2020improved}. 

With the emergence of machine learning, various learning-based approaches~\cite{sun2019test,millican2019applying, ma2019high} achieve comparable test results to the previous work. \cite{sun2019test} predicts the test coverage improvement of different TPs by an artificial neural network (ANN). After that, \cite{millican2019applying} estimates the TP impact on delay fault coverage improvement. However, the label generation relies on the exhaustive fault simulation after inserting all possible TPs, which is expensive and intractable for large-scale circuits. \cite{ma2019high} proposes a graph convolutional network (GCN) to classify nodes as either hard-to-observe or easy-to-observe. The application of GCN to circuits is shown to be effective for automatic test pattern generation (ATPG) TPI.
Nevertheless, the labels used for model training may not be always accurate because they are obtained from commercial DFT tools, which in turn adopt heuristic algorithms. In theory, the model performance is bounded by the heuristic algorithm and is sub-optimal. To summarize, the supervised learning approach is not fully suitable for the TPI problem.

Instead of collecting supervision for all TP candidates from the DFT tools, in this work, we propose a novel learning-based TPI technique, namely \textit{DeepTPI}, which formulates the TPI problem as a reinforcement learning (RL) task. The optimization objective of DeepTPI is to maximize LBIST test coverage by inserting TPs sequentially. Compared with the prior supervised-learning based TPI approaches, the benefits of our proposed solution are as follows: 1) The RL-based approach is more sample-efficient than supervised learning and is free from the dependence on the vast number of training samples; 2) The RL-based approach can fundamentally break the upper bound of performance caused by supervision. 

In the DeepTPI formulation, we transform the circuit netlist into a graph to naturally represent the circuit. The RL action space is defined as inserting a TP following a certain gate over the circuit graph. With the node-level discrete action space, our RL agent is trained with Deep Q-learning~\cite{mnih2015human}, and is equipped with a novel graph-based deep Q-network (\textit{Graph-DQN}) to fit the action-value function. After aggregating the message from neighbors for multiple iterations, the hidden state of each node is read out to predict the expected total reward of all possible actions for this node. We set the test coverage improvement as the reward in this TPI-for-LBIST task. The RL policy is defined as selecting the action with the maximum expected value.
We also enhance the representation ability of our \textit{Graph-DQN} by involving prior knowledge from DeepGate~\cite{li2021representation}. 
In addition, we analyze the dominant factors of testability and mimic testability by a customized attention mechanism. During GNN aggregation, the testability-aware attention mechanism gives higher weight to those nodes with more significant impacts on testability. 

The contributions of this work are summarized as follows: 
\begin{itemize}
    \item We formulate the TPI problem as a reinforcement learning task and train an RL agent with the Deep Q-learning algorithm. To the best of our knowledge, this is the first work applying reinforcement learning on test point insertion. 
    \item We transform the RL state into a graph and deploy a dedicated graph-based deep Q-network (Graph-DQN) to assist RL decisions. Moreover, the general neural representation from pre-trained DeepGate~\cite{li2021representation} is embedded into the node feature to involve the prior circuit knowledge. 
    \item We deploy a dedicated GNN to assist RL policy decisions. By analyzing the essence of testability, we propose a testability-aware attention mechanism in the GNN to capture more useful information during aggregation for the TPI task. 
\end{itemize}

We organize the remaining sections as follows. Some related work about this paper is summarized in Section~\ref{Sec:Related}. Section~\ref{Sec:RL} models TPI as an RL problem and introduces the RL components. The architecture and methodology of DeepTPI are explained in Section~\ref{Sec:Method}. The experimental results are shown in Section~\ref{Sec:Experiment}. Finally, Section~\ref{Sec:Conclusion} concludes this paper.

\section{Related Work} \label{Sec:Related}

\subsection{Test Point Insertion}\label{Subsec:TPI}

\begin{figure}[t!]
	\centering
	\includegraphics[width=\linewidth]{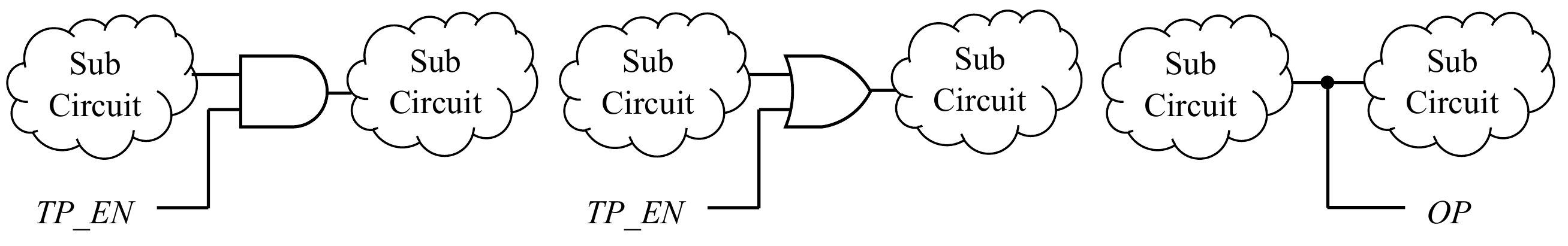}
	\caption{Examples of Test Point Insertion(Left: AND Control Point, Middle: OR Control Point, Right: Observation Point)}
	\label{FIG:TPI_TP}
\end{figure}

Test point insertion~\cite{williams1973enhancing} is a circuit modification technique that adds extra gates into circuits during the design-for-test (DFT) phase and enhances circuit testability. As shown in Fig.~\ref{FIG:TPI_TP}, the test points are categorized into two types: control points (CPs) and observation points (OPs). CPs are typically implemented as OR gates for control-1 TPs or AND gates for control-0 TPs. During the testing, a test enable pin forces lines to their controlled values~\cite{yang2011test}. The goal of control CPs is to increase the circuit \textit{controllability}, i.e., the probability of activating faults and sensitizing the fault propagation paths~\cite{sun2021novel}. 
OPs change circuit \textit{observability} by inserting pseudo primary output to the gate outputs to make faulty values on lines more observable~\cite{fox1977test}.

Although TPI has various applications in different scenarios, e.g., ATPG~\cite{acero2015embedded} or analog testing~\cite{zhang1999test}, this work addresses TPI in the context of LBIST. 
On the one hand, LBIST generates test patterns and verifies the corresponding responses on-chip. Thus, as an in-field test technique, LBIST is in demand for high-assurance applications, such as implantable medical devices and self-driving vehicles. 
On the other hand, LBIST quality is more suffered from the presence of RPR faults than the deterministic testing~\cite{mccluskey1985built}. Handling the RPR faults by TPI is known as a \textit{NP-hard} problem~\cite{krishnamurthy1987dynamic}. The current heuristic TPI approaches are still inefficient~\cite{he2017test} and even decrease the circuit testability in some cases~\cite{roy2020improved}. 

Since the performance of most conventional TPI approaches degrades for the modern circuits with complex structures, various learning-based approaches are proposed to select TPs using advanced machine learning algorithms. \cite{sun2019test} and \cite{millican2019applying} estimate the quality of TPs with an artificial neural network (ANN), but the model relies on large amounts of samples. Unlike supervised learning approaches, the RL algorithm used in this paper trains an agent by action-environment interaction rather than exhaustively executing all possible TP candidates. \cite{ma2019high} classifies each node into easy-to-observe or hard-to-observe. However, the binary labels are collected from a heuristic testability analysis algorithm, which is sub-optimal and bounds the performance of the resulted model in theory. On the contrary, our proposed RL-based approach is not limited by the upper-bound performance.

\subsection{Reinforcement Learning}

Recently, reinforcement learning (RL) has achieved outstanding performance in different applications. In general, RL is suitable for solving the problems without apparent supervision, which is often the case in many EDA applications. For example, getting the optimal test patterns in software-based self-test (SBST) is impractical. Thus, \cite{DBLP:conf/ats/ChenH19} formulates SBST as an RL problem and rewards the agent when a newly detected fault is reached. Besides, the best quality of result (QoR) is hard to be produced as supervision in logic synthesis due to the exponential number of possible optimization permutations. \cite{8351885, 9045559} convert logic synthesis as an RL problem, whose optimization target is set as reducing area or depth. 

In EDA, the most natural representation of circuits, intermediate RTL, netlists, and layouts are graphs~\cite{lopera2021survey}.
Consequently, a few RL approaches incorporate graph neural networks as the backbones to capture the graph structures. For instance, GCN-RL circuit designer~\cite{wang2020gcn} embeds the topology graph information with GNN and uses an RL optimization algorithm to achieve the high Figures of Merit (FoM) for automatic transistor sizing problems. \cite{DBLP:journals/corr/abs-2004-10746} proposes a GNN model to encode the information about netlist and make macro placement decisions with another policy network. 
Besides, \cite{khalil2017learning} and \cite{kurin2020can} combine the RL agent with GNN to design combinatorial optimization solvers. Our proposed DeepTPI shares the similar intuition with these works, which learns the circuit knowledge via a GNN and conducts TPI actions based on the learned graph embeddings.

\section{Reinforcement Learning Formulation} \label{Sec:RL}

\begin{figure}[t!]
	\centering
	\includegraphics[width=0.8\linewidth]{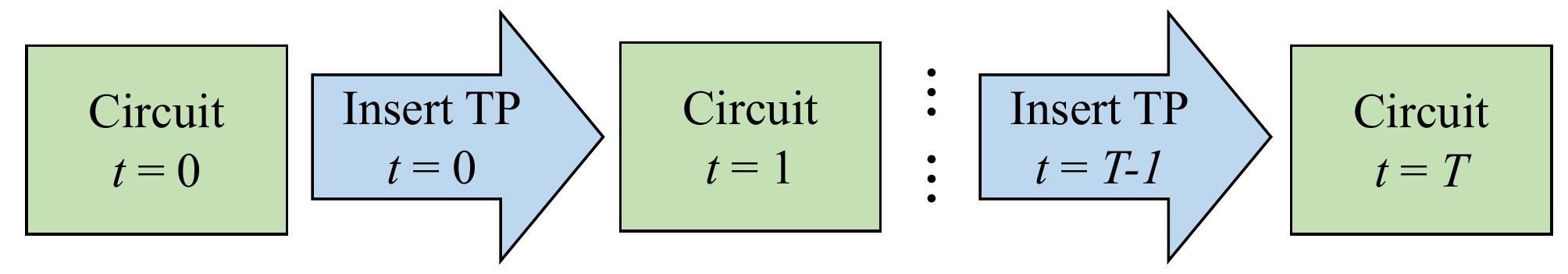}
	\caption{Test Point Insertion as an Markov Decision Process (MDP)}
	\label{FIG:MDP}
\end{figure}

In this work, we target the test point insertion (TPI) problem, in which the objective is to improve the test coverage under LBIST random patterns by inserting control points and/or observation points.
Such a \textit{sequential decision-making} problem can be modeled as a Markov Decision Process (MDP). Moreover, an optimal decision strategy in such a sequential decision-making setting can be identified by reinforcement learning (RL) methods~\cite{clifton2020q}. Hence, in this section, we formally introduce the TPI problem from a reinforcement learning perspective, and build the connections between key elements of RL and TPI.

Given a circuit netlist, we form a directed graph $\mathcal{G}=(\mathcal{V}, \mathcal{E})$, where each gate $v$ in the circuit is a vertex belong to $\mathcal{V}$, and $\mathbf{e}(u,v) \in \mathcal{E}$ is a directed edge from node $u$ to node $v$. We use $T\in \mathbb{N}$ to denote the time horizon, i.e., the number of test points to be inserted. The circuit after inserting $t$ test points can be represented as $\mathcal{G}^t$. Equipped with the above notations, we then formulate the TPI problem as an MDP (Shown in Fig.~\ref{FIG:MDP}), which consists of the following four key elements:

\begin{itemize}
\item State $\mathcal{ST}$: the set of states. In TPI task, the state is reflected as circuit with graph structure. The state is represented as the circuit at time step $t$: $s^t := \mathcal{G}^{t} \in \mathcal{ST}$.

\item Action $\mathcal{A}$: the set of actions that an agent can execute, i.e., inserting test points into the circuit. The action corresponds to inserting one kind of TPs on the following wire of a particular gate, which can be defined as $a^t := (v, type) \in \mathcal{A}$, wherein $v$ is the selected gate for the test point insertion and $type \in \{\text{AND-CP}, \text{OR-CP}, \text{OP}\}$ indicates what type of test point to be inserted. 

\item State transition function $\mathcal{F}$: Given state $s^{t}$ and an action $a^{t}$ at time $t$, the function produces the updated state $s^{t+1}$. Since the the circuit after an action of TPI is uniquely determined, the state transition function is deterministic. We denote the transition function as $s^{t+1} = \mathcal{F}(s^{t}, a^{t})$. 

\item Reward function $\mathcal{R}$: step reward $r^t = \mathcal{R}(s^t, a^t, s^{t+1})$ caused by the corresponding action $a^t$ with state transition from $s^t$ to $s^{t+1}$. In TPI task, we consider the improvement of the test coverage as the reward.

\end{itemize}

\begin{figure}[t!]
	\centering
	\includegraphics[width=0.8\linewidth]{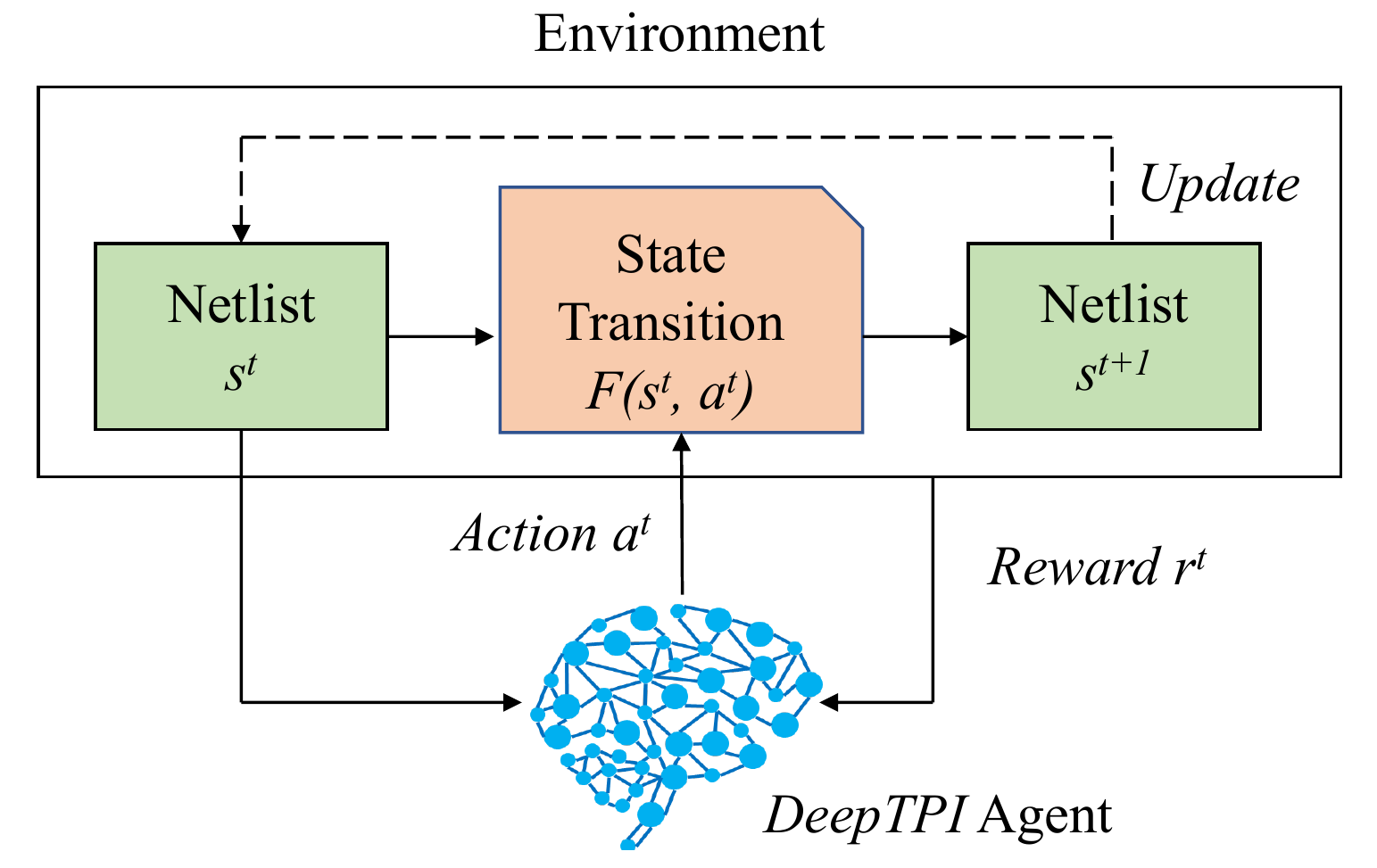}
	\caption{RL Agent - Environment Interaction in DeepTPI}
	\label{FIG:Agent}
\end{figure}

As shown in Fig.~\ref{FIG:Agent}, at each step $t$, the agent extracts information from the current state $s^t$ and selects an action $a^t$. Then, the circuit inside the environment is updated by the state transition function $s^{t+1} = \mathcal{F}(s^{t}, a^{t})$. The primary target to solve the MDP problem is to train an RL agent with optimal policy $\pi$, a mapping that outputs an action which maximizes the expected cumulative rewards. The MDP also defines a discounting factor $\gamma^0$, which determine the different weights on long-term reward and short-term reward. The overall objective of the agent decision is to maximize the expected discounted reward: $G=\mathbb{E}[\sum_{t=0}^{T-1} \gamma^t r^t]$, where $\gamma^t = \text{Power(}\gamma^0, t\text{)}$. 

In the TPI setting, the trained agent shall select the optimal positions and insert the corresponding test points so that the maximum test coverage improvement is achieved. We instantiate the reward function $\mathcal{R}(s^t, a^t, s^{t+1})$ in Eq.~\eqref{FML:Reward} as test coverage improvement (\textit{TC Imp.}) after inserting $T$ TPs in total, that is the difference of test coverage between the final circuit with $T$ TPs inserted and the initial circuit. The reward function belongs to the \textit{terminal} reward, which is nonzero only after inserting $T$ TPs. 
Thus, we can set the discounting factor $\gamma$ to an arbitrary value in $[0, 1)$ because the long-term or short-term discounting does not affect the optimization objective. 
Ideally, the trained RL agent selects a list of TPs with the size of $T$ to achieve globally optimal test coverage improvement. 

\begin{equation} \label{FML:Reward}
    r^t =  \mathcal{R}(s^t, a^t, s^{t+1}) = \left\{
        \begin{aligned}
            & 0, \qquad \qquad t = 0,\cdots,T-2, \\
            & TC\text{ }Imp., \ t = T-1
        \end{aligned}
            \right.
\end{equation}

To approach the described objective above, we designate a combination of graph neural network and deep Q-learning that approximates an optimal action-value function for TPI. We discuss the methodology in detail in the following section.

\section{Methodology} \label{Sec:Method}
\begin{figure*}[!t]
	\centering
	\includegraphics[width=\linewidth]{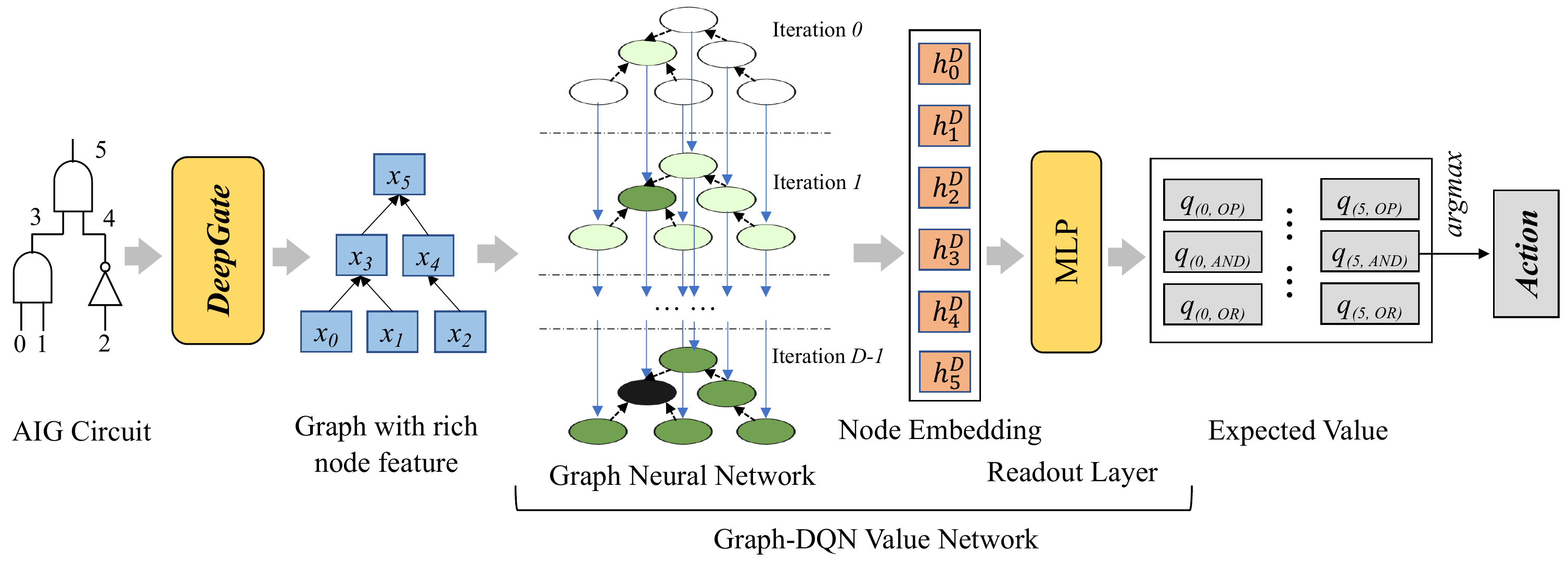}
	\caption{The Overview of DeepTPI Agent}
	\label{FIG:overview}
\end{figure*}

\subsection{Overview of DeepTPI}
Fig.~\ref{FIG:overview} shows the overview of the DeepTPI RL agent.
Firstly, the input circuits are preprocessed into a unified format along with the rich node features extracted from a pre-trained model (i.e., DeepGate~\cite{li2021representation}). More precisely, given a pool of circuit netlists for training, we process them via an equivalent transformation into And-Inverter Graph (AIG), wherein only three gate types, i.e., primary input (PI), \textit{AND} and \textit{NOT}. The details of such transformation are elaborated in Sec.~\ref{Sec:Method:AIG}. Besides the circuit transformation, we present a \textit{feature-based} approach to enrich the node features. We extract the node embeddings for each gate from a pre-trained DeepGate model, which learns the general-purpose gate embeddings (See Sec.~\ref{Sec:Method:pretrain}).

Secondly, since the TPI procedure can be formulated as an MDP with a discrete action space as discussed in Sec.~\ref{Sec:RL}, the value-based deep Q-learning~\cite{mnih2015human} algorithm can be applied for this task. We design a Graph-based Deep Q-Network (\textit{Graph-DQN}), instantiated as a GNN along with a readout layer. The Graph-DQN consumes the input graphs and learns to capture the testability properties of a logic gate in the context of the whole circuit (See Sec.~\ref{Sec:Method:GNN}). To better embed testability by the GNN model, we analyze the requirements of faults being detectable and 
propose a novel testability-aware attention mechanism in the aggregation function. The customized attention-based aggregation function enables differentiation of predecessors and successors according to their contribution to fault sensitization and fault propagation, respectively (See Sec.~\ref{Sec:Method:Attention}).

Finally, given expected values for all actions predicted by Graph-DQN, the RL agent selects the action with the maximum expected value to be conducted. Such pipeline is repeated under the interaction with the environment until a pre-defined termination criterion is satisfied (e.g., hit the maximum number of test points to be inserted). Alg.~\ref{ALGO:Overview} provides the pseudo-code of DeepTPI. 

\begin{algorithm}[t]
\caption{DeepTPI Application Process}
\label{ALGO:Overview}
\KwIn{The pre-TPI circuit $ckt$, Trained value network \text{Graph-DQN} and Pre-trained model \text{DeepGate}}
\KwOut{The post-TPI circuit $ckt^{'}$}
    \tcc{Initialize circuit graph $\mathcal{G} = (\mathcal{V}, \mathcal{E})$}
    $\mathcal{G}^0 = \text{ConvertAIGGraph(} ckt \text{)}$\;
    \tcc{RL decision}
    Define TP list $\ell$\;
    Define time-step $t = 0$\;
    
    \While{$t \leq \text{number of TPs}$}{
        \tcc{Calculate the initial feature}
        $node\_emb^t = \text{DeepGate}(\mathcal{G}^t)$\;
        $\mathbf{X}^t$ = \text{Concatenate}($node\_emb^t$, OneHot($\mathcal{V}^t$)) \;
        
        \tcc{Calculate expected value}
        $\mathbf{Q}^t = \text{Graph-DQN}(\mathcal{G}^t, \mathbf{X}^t)$\;
        
        \tcc{Choose action}
        $a^t = \text{GetAction(} \arg \max(\mathbf{Q}^t) \text{)}$\;
        $\ell$.append($a^t$)\;
        
        \tcc{Transit state}
        $\mathcal{G}^{t+1} = F(\mathcal{G}^t, a^t)$\;
    }
    
    \tcc{Generate the post-TPI circuit}
    $ckt^{'} = \text{InsertTPs(} ckt, \ell \text{)}$\;
    \Return $ckt^{'}$\;
\end{algorithm}

\subsection{Unified AIG of Circuits} \label{Sec:Method:AIG}
\begin{figure}[!t]
	\centering
	\includegraphics[width=0.4\textwidth]{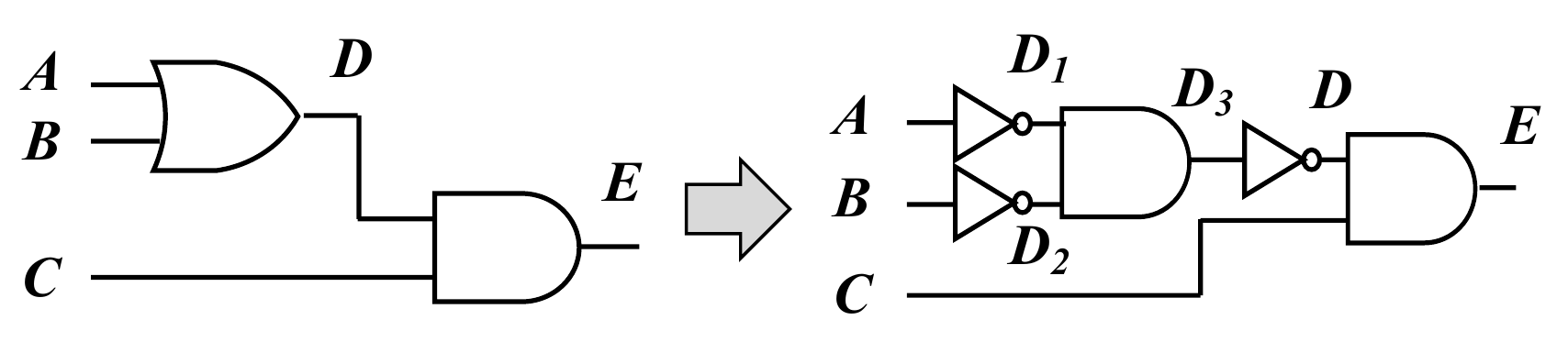}
	\caption{Convert Circuit to AIG}
	\label{FIG:AIG}
\end{figure}

We opt for a \textit{unified AIG} representation of circuits.
Since the different circuits in the DFT phase are mapped into standard cells from different libraries, the gate types and local structures of the circuits are likely to have non-uniform distribution. Such heterogeneity across circuit designs is a challenge for GNN and RL model development. Taking the fact that any type of logic gate can be converted into the combination of \textit{AND} gate and \textit{Inverter} (\textit{NOT} gate), we map circuits to their corresponding AIG counterparts. It should be noted that the transformation process does not rely on the AIG synthesis tool~\cite{brayton2010abc} because test point insertion in the DFT stage prohibits modifying the circuit structure except for TPs. Instead, we simply map the original gates to their equivalent And-Inverter combinations without logic optimization. 
Moreover, only those gates in AIG that have the corresponding positions in original circuits are considered TP candidates. Fig. ~\ref{FIG:AIG} shows a toy example of the applied transformation. Only $A, B, C, D, E$ are considered as the candidate positions for the RL agent. $D_1, D_2, D_3$ are masked out because these nodes in AIG format do not correspond to any gates in the original circuit.

There are two benefits of the AIG transformation: 
1) we map the gate into the combination of \textit{AND} and \textit{NOT} gates directly rather than optimizing the logic so that the transformation is of linear complexity and can scale on large circuits;
2) including PI, only three types of nodes are considered, which would dramatically reduce the difficulty of RL training.

\subsection{Features of Logic Gates from DeepGate}\label{Sec:Method:pretrain}
A natural way to assign the node features for logic gates is to use the one-hot encoding of gate types. When PIs are taken into account, there are three gate types in AIG: $PI$, \textit{AND} and \textit{NOT}. Even though such one-hot encoding is a widely adopted convention for graph learning~\cite{thost2021directed, abs-1904-11088}, it cannot include any specific properties of logic gates to benefit RL decision, such as the gate behaviors and testability. With limited information on logic gates to utilize, the RL model may struggle to learn and exposes the RL agent to a high risk of corruption. 

To augment the semantic information embedded in node features, we follow the success of model pre-training techniques in deep learning community~\cite{devlin2018bert, brown2020language}, and adopt a \textit{feature-based} approach, i.e., using the pre-trained node embeddings as the additional features to incorporate the gate semantics. To be specific, we extract the node embeddings from a pre-trained DeepGate model~\cite{li2021representation} and feed them as the initial node features for RL training. The task of DeepGate is to approximate the simulated probability, i.e., the probability of the gate being logic $1$ under random logic simulation. With strong biases introduced into DeepGate design, the node embeddings for logic gates embed both logic function and structural information of a circuit. We concatenate the node embeddings from pre-trained DeepGate and the $3$-d one-hot encoding of gate types together to construct the node feature $x_i$ for node $v_i$. The effectiveness of the proposed feature-based pre-training is demonstrated by the ablation study in Sec.~\ref{SEC:EXP:pretrain}.

\subsection{Graph-based Deep Q-Network} \label{Sec:Method:GNN}

In the MDP formulation described in Sec.~\ref{Sec:RL}, we define the action as inserting a TP following candidate gate, and the state transition as modifying the circuit with the newly inserted TP.
The Q-learning algorithm is suitable to explore an optimal policy for such MDP problem with a \textit{discrete} gate-level action space and \textit{deterministic} state transition~\cite{watkins1992q}. Therefore, we  apply Q-learning to RL-formulated TPI problem. 

The classical Q-learning approximates the expected reward for all possible actions based on the current state, and maintains an action-value function as a table (also named Q-Table). Due to the increasing scale of action space in practice, the deep Q-learning~\cite{mnih2015human} introduces a value network (also named Q-Network) to implicitly fit the action-value function. Specifically, at each time step $t$, the Q-Network ($\mathcal{Q}$) predicts the future rewards of all possible actions $a^t$ in current state $s^t$. The optimal policy $\pi$ is defined as executing the action with maximum value:
\begin{equation}
    a^t = \pi(s^t) = \arg \max_{a^t}(\mathcal{Q}(s^t, a^t))
\end{equation}

In TPI task, since the state $s^t$ solely depends on a directed acyclic graph $\mathcal{G} = (\mathcal{V}, \mathcal{E})$, the action space is defined at node-level over an arbitrary graph. To accommodate the requirements of dealing with graph-based data, we refer to a graph neural network, and implement the value network as the graph-based deep Q-Network (\text{Graph-DQN}) to approximate the action value.  
The whole architecture is shown as the \text{Graph-DQN} value network block in Fig.~\ref{FIG:overview}. 
We calculate the node embedding for each node for multiple iterations by a GNN and readout the expected value with a multi-layer perceptron (MLP).

We formally introduce the GNN structure used in \text{Graph-DQN}. The GNN propagates the message between each node $v_i \in \mathcal{V}$ and its neighbors $\mathcal{N}(i)$ with two functions: aggregation function \text{AGGR} and update function \text{UPDATE}. The aggregation function enables the message-passing of all nodes with arbitrary degrees. The update function combines the node feature with the incoming message and updates the hidden state iteratively. We apply gated recurrent neural (GRU)~\cite{chung2014empirical} to implement the \text{UPDATE} function. 
We define the GNN computations as Eq.~\ref{FML:GNN:AGGE}, where $h_i^d$ is the hidden state and $d = 0, 1, ..., D-1$ is the number of iterations. The initial hidden state $h_i^0$ is assigned to node feature $x_i$. The aggregation function can be implemented by average pooling~\cite{hamilton2017inductive} or linear transformation~\cite{kipf2016semi}, but these methods do not differentiate the neighbors who dominate the testability. To mimic the essence of testability, we borrow the attention-based aggregation function from~\cite{velivckovic2017graph}. We elaborate on the testability-aware attention mechanism in Sec.~\ref{Sec:Method:Attention}.

\begin{equation} \label{FML:GNN:AGGE}
    h_i^{d+1} = \text{UPDATE}^{d+1}(\text{AGGR}^{d+1}(\{h_k^{d+1}\| k \in \mathcal{N}(i)\}), h_i^{d})
\end{equation}

The GNN message-passing process performs for $D$ iterations. The $h_i^{D}$ after the node embedding of the final iteration $d=D-1$ is fed into the readout function $\text{READOUT}$ to produce the expected total value of all possible actions:
\begin{equation} \label{FML:GNN:READOUT}
    q_i = \text{READOUT}(h_i^D)
\end{equation}
where $q_i$ is a three-dimension vector for each node $v_i \in \mathcal{V}$ that presents the action value when inserting the AND control point, OR control point, or observation point. The RL agent chooses the action with the maximal value to maximize the total reward.

We define the action selection process as following four steps:
\begin{enumerate}
    \item \textit{Node feature construction}: 
    The initial node feature $\mathbf{X}^t$ is concatenated of the node embedding $node\_emb^t$ from the pre-trained DeepGate model and one-hot gate type encoding of each node $\text{OneHot}(\mathcal{V}^t)$. 
    \begin{equation}
        \begin{split}
            node\_emb^t & = \text{DeepGate}(\mathcal{G}^t) \\
            \mathbf{X}^t & = \text{Concatenate}(node\_emb^t, \text{OneHot}(\mathcal{V}^t))
        \end{split}
    \end{equation}
    
    \item \textit{Value approximation}: 
    Based on the initial node feature $\mathbf{X}^t$ and graph $\mathcal{G}^t$, the proposed Graph-DQN predicts the expected cumulative reward $\mathbf{Q}^t$ for each possible action in node-level. 
    \begin{equation}
        \mathbf{Q}^t = \text{Graph-DQN}(\mathcal{G}^t, \mathbf{X}^t)
    \end{equation}
    
    \item \textit{Action execution}: The agent policy is selecting the action with the maximum expected reward.
    \begin{equation}
        a^t = \text{GetAction}(\arg \max(\mathbf{Q}^t))
    \end{equation}
    
    \item \textit{State transition}: The circuit structure changes after inserting the corresponding TP. The environment updates the state to be $\mathcal{G}^{t+1}$. 
    \begin{equation}
        \mathcal{G}^{t+1} = F(\mathcal{G}^t, a^t)
    \end{equation}
\end{enumerate}

Besides, in the RL training process, the parameter $\theta$ in $\text{Graph-DQN}_{\theta}$ is updated iteratively by:
\begin{equation}
    \begin{split}
        L(\theta) & = Loss (\text{Graph-DQN}_{\theta}(s^t, a^t), 
                       \mathcal{R}(s^t, a^t, s^{t+1}) + \\ & \gamma^0  argmax_{a^{t+1}}(\text{Graph-DQN}_{\hat{\theta}}^{'}(s^{t+1}, a^{t+1})))
    \end{split}
\end{equation}
where the target network $\text{Graph-DQN}_{\hat{\theta}}^{'}$ is another value network used as a measure of ground truth (i.e. the expected future reward) and stabilize learning. The parameter $\hat{\theta}$ is updated to $\theta$ every $U$ epochs. We use Mean-Squared (MSE) loss as the loss function $Loss$.

\subsection{Testability-Aware Attention Mechanism} \label{Sec:Method:Attention}
\begin{figure}[!t]
	\centering
	\includegraphics[width=0.3 \textwidth]{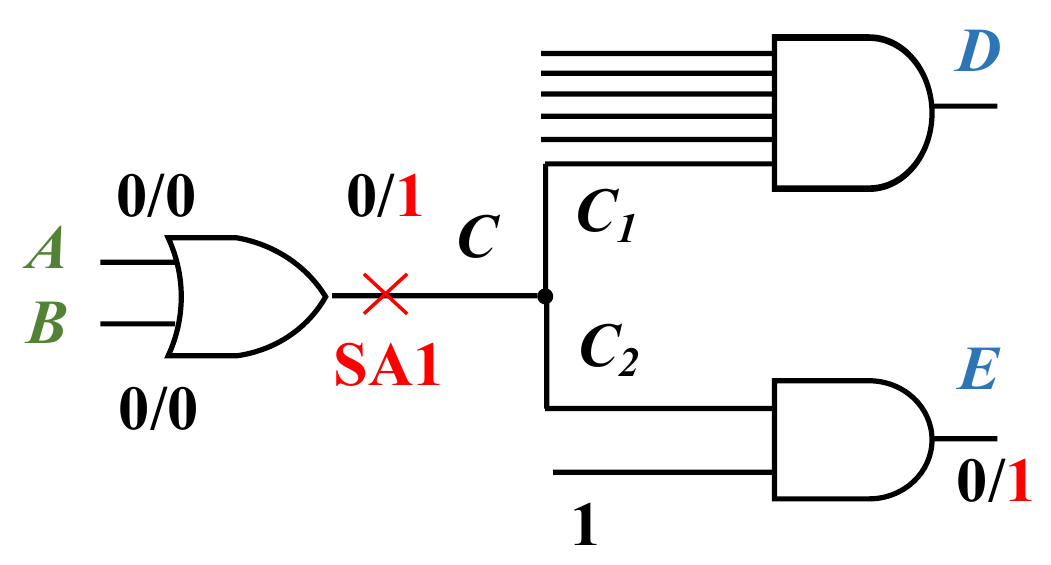}
	\caption{An Example of Stuck-at-1 Fault}
	\label{FIG:Attn}
\end{figure}

In the logic circuits, when the logic gate receives a \textit{controlling value} from one of the inputs, the output value is almost independent of the other inputs~\cite{devadas1994certified}. For example, the fan-in value 0 is a controlling value of AND gate and forces the gate to be 0 no matter what the values of the other fan-in gates are. The rule can be derived from the other types of gates.
Such similar gate property plays an important role in testability analysis. Fig.~\ref{FIG:Attn} shows an example of detecting a stuck-at-1 (SA1) fault at gate $C$, where 0/1 means the expected logic value of this line is 0, but the fault causes the value to 1. The circuit must fulfill two conditions to enable such SA1 fault to be detected: (a) at least one random pattern assigning logic 0 at the gate $C$, i.e., the fault-effect (logic difference) is activated; (b) at least one path that allows the fault-effect to propagate from $C$ to any primary outputs $E$ or $D$, i.e., the fault-effect can be observed. We assume the testability as the probability of a gate meeting both conditions (a) and (b).

The predecessor with controlling value dominates the fault activation. For example, if there is a controlling value (logic 1) in any fan-in wire of gate $C$, the fault cannot be activated. Intuitively, to estimate the probability of meeting the above condition (a), we need focus on the fan-in gates with a high probability to be logic 1. Moreover, condition (b) requires the fault-effect being observed, no matter on which primary output. Take the fault-effect 0/1 propagation in Fig.~\ref{FIG:Attn} as an example, there are two paths following stem gate $C$. Sensitizing the path through $E$ by forcing one line to be 1 is easier than $D$. The probability of observing this fault-effect relies only on the difficulty of sensitizing the path through successor $E$. Therefore, the neighbor that can easily propagate fault-effect is more noteworthy. To summarize, the predecessors and successors play different and significant roles in the testability of each gate. 

To mimic the above properties during GNN propagation, we turn to graph attention network~\cite{velivckovic2017graph}. Graph attention network computes the hidden representations of each node in the graph by attending over its neighbors following a attention mechanism. Inspired by this work, our RL value network also follows an attention strategy. The Graph-DQN distinguishes the attention on predecessors and successors because the predecessors determine whether the fault-effect can be activated, and the successors affect the effect observation. Therefore, we define the testability-aware attention mechanism in the aggregation function $\text{AGGR}$ as below:
\begin{equation} \label{FML:ATTR}
    \begin{split}
        m_{i(pre)}^d & = \sum_{p \in \mathcal{P}(i)} \alpha_{(i, p)}^d h_p^d, \\ 
        & \ \text{where}\ \alpha_{(i, p)}^d = \mathop{softmax}\limits_{p \in \mathcal{P}(i)}(w_0^{\top} h_i^{d-1} + w_{pre}^{\top} h_p^d) \\
        m_{i(suc)}^d & = \sum_{s \in \mathcal{S}(i)} \beta_{(i, s)}^d h_s^d, \\ 
        & \ \text{where}\ \beta_{(i, s)}^d = \mathop{softmax}\limits_{s \in \mathcal{S}(i)}(w_0^{\top} h_i^{d-1} + w_{suc}^{\top} h_s^d)
    \end{split}
\end{equation}
wherein $\mathcal{P}(i)$ and $\mathcal{S}(i)$ are the predecessors and successors of node $i$, respectively. During the training process, two independent model parameters, $w_{pre}$ and $w_{suc}$, are used for calculating the attention weighting coefficients $\alpha_{(i, p)}^d$ and $\beta_{(i, s)}^d$. These two coefficients represent the appropriate attention on aggregating information from different predecessors or successors.

After we obtain the aggregated message $m_{i(pre)}^d$ and $m_{i(suc)}^d$, we update them together as the output of $AGGR$ function. With the testability-aware attention mechanism, the GNN can generate the embedding with much more testability information for RL decisions.


\section{Experiments} \label{Sec:Experiment}
In this section, we present our constructed circuit datasets for training and testing, as well as the baseline in Sec.~\ref{SEC:EXP:dataset}. We elaborate the model settings in Sec.~\ref{SEC:EXP:setting}. Then, we discuss the performance comparison between DeepTPI and the commercial DFT tool, and analyze the advantage of DeepTPI in Sec.~\ref{SEC:EXP:com}. To explore the effectiveness of the testability-aware attention mechanism and the feature-based pre-training, we present two ablation studies in Sec.~\ref{SEC:EXP:Attn} and Sec.~\ref{SEC:EXP:pretrain},  respectively. Finally, we analyze the computational complexity and discuss some potential improvements in Sec.~\ref{SEC:EXP:Comp}.

\subsection{Circuit Dataset and Baseline}\label{SEC:EXP:dataset}
We extract 400 sub-circuits from ISCAS89~\cite{ISCAS89} and ITC99~\cite{ITC99}, and convert them to AIG format for RL training. Tab.~\ref{TAB:Train} summarizes the statistics of training circuits. \#~Gates denotes the number of gates in circuits, while \# PIs and \# POs represent the number of primary inputs (PIs) and primary outputs (POs), respectively. \#~Levels is the total number of logic levels. The test coverage (TC) on original circuits is reported by a commercial LBIST tool under the same volume of random test patterns, i.e., $300K$ patterns.
In some cases, especially for these circuits with high test coverage, inserting the inappropriate test point may reduce the random test coverage. For example, the extra AND control point on the fan-in wires of the AND gates causes the SA1 faults more hard to be activated. To explore the optimal policy avoiding the test coverage loss for these circuits, our training dataset also includes the circuits that can be fully tested by LBIST (TC=100.00\%). 

\begin{table}[!t]
\centering
\caption{The Statistics of Training Circuits} \label{TAB:Train}
\begin{tabular}{@{}llllll@{}}
\toprule
     & \# Gates & \# PIs & \# POs & \# Levels & TC       \\ \midrule
Min. & 83       & 8      & 1      & 7         & 24.08\%  \\
Max. & 656      & 98     & 24     & 80        & 100.00\% \\
Avg. & 201.73   & 23.87  & 7.42   & 15.77     & 90.59\%  \\ \bottomrule
\end{tabular}
\end{table}

For inference, we choose eight large combinational circuits from two open-source benchmarks, i.e.,  ITC~\cite{ITC99} and EPFL~\cite{EPFLBenchmarks}, and compare the test coverage reported from the state-of-the-art commercial DFT tool and DeepTPI. To examine the generalization property of our RL approach, the size of testing circuits varies from 3.4K to 127.4K, which is much larger than those in the training dataset. The statistics of the testing circuits in AIG format are listed in Tab.~\ref{TAB:Stat}. It should be noted that there is no overlap between training circuits and testing circuits. The pre-TPI test coverage is shown in Column TC in Tab.~\ref{TAB:Stat}. 

According to practical experience, the inserted test points for LBIST coverage improvement occupy 0.1\%-2.0\% extra area over the entire circuit. Typically, there is a strong positive correlation between the number of TPs and test coverage improvement. In the following experiments, we fix the number of inserted test points (\# TPs) as $1.0\%$ of \# Gates for all testing circuits. 

\begin{table}[!t]
\renewcommand\tabcolsep{2.5pt}
\centering
\caption{The Statistics of Testing Circuits in AIG format} \label{TAB:Stat}
\begin{tabular}{@{}lllllll@{}}
\toprule
   & Name      & \# Gates & \# PIs & \# POs & \# Levels & TC \\ \midrule
D1 & b15\_C    & 27,838   & 485    & 449    & 158   & 90.61\% \\
D2 & b20\_C    & 61,667   & 522    & 507    & 165   & 90.60\%  \\
D3 & b21\_C    & 62,817   & 522    & 507    & 167    & 89.73\%    \\
D4 & b22\_C    & 91,505   & 767    & 750    & 173    & 91.70\% \\
D5 & i2c       & 3,400    & 136    & 127    & 34   & 86.11\%       \\
D6 & max       & 9,439    & 512    & 129    & 333   & 52.65\%     \\
D7 & b17\_C    & 99,466   & 1,452   & 1443   & 234   & 86.84\%   \\
D8 & mem\_ctrl & 127,353  & 1,028   & 967    & 195    & 69.95\%    \\ \bottomrule
\end{tabular}
\end{table}

\subsection{Model Settings} \label{SEC:EXP:setting}
In the Graph-DQN model configuration, we set the dimension of node hidden state as 64 and the iteration of message passing as 10. We also involve the universal logic gate representation pre-trained model DeepGate. The pre-trained DeepGate keeps the same configurations and hyperparameters as the original specification~\cite{li2021representation}. We set the dimension of node hidden state as 64 and the iteration of hidden state update as 10 for DeepGate. The DeepGate model is trained with 10,824 subcircuits extracted from open-source benchmarks. 

\begin{table}[!t]
\caption{The Model Complexity of \text{Graph-DQN} and DeepGate} \label{Tab:FLOPs}
\centering
\begin{tabular}{@{}llll@{}}
\toprule
           & \text{Graph-DQN} & DeepGate & Total    \\ \midrule
\# FLOPs  & 116.47 M & 60.63 M  & 177.10 M \\
\# Param.  & 59.01 K  & 69.13 K  & 128.14 K \\ \bottomrule
\end{tabular}
\end{table}

The model complexity is illustrated in Tab.~\ref{Tab:FLOPs}, where \text{\# FLOPs} is the number of floating-point operations and \text{\# Param.} is the total number of trainable parameters. The RL agent is trained for 500 episodes on each circuit in Tab.~\ref{TAB:Stat}. We use the ADAM optimizer with the learning rate $lr=0.0001$. In the following experiment, we keep the same configuration for all DeepTPI agents.

\subsection{Performance Comparison with Commercial DFT Tool} \label{SEC:EXP:com} 

This section compares the TPI performance between the state-of-the-art commercial DFT tool and our proposed DeepTPI. The user manual of the tool indicates that the testability analysis algorithm is based on COP~\cite{COP1984OnTO} and pattern simulation~\cite{schotten1995test}. 

To ensure fairness, we use the same pre-TPI circuits and assign the same number of inserted TPs. The random test patterns are generated by the same commercial LBIST tool with an equivalent configuration. One group of post-TPI circuits is produced by the commercial tool and the other is generated by sequential actions from DeepTPI. The test coverage of these two groups are abbreviated as Com. Tool TC. and DeepTPI TC, respectively. As shown in Tab.~\ref{TAB:comm}, DeepTPI outperforms the commercial DFT tool significantly. For instance, our agent can achieve $4.78\%$ test coverage improvement (abbreviated as Imp.) on the testing circuit D3 higher than that $0.53\%$ from the commercial DFT tool, which reaches $9.02\times$ improvement. On average, we can achieve $2.95\times$ test coverage improvement than the tool. 



\begin{table}[!t]
\renewcommand\tabcolsep{2.5pt}
\centering
\caption{The Test Coverage Improvement Comparison between DeepTPI and Commercial DFT Tool} \label{TAB:comm}
\begin{tabular}{@{}lcc|cc|cc@{}}
\toprule
              & TC        & \# TPs       & Com. Tool TC & Imp.           & DeepTPI TC & Imp.           \\ \midrule
D1            & 90.61\%   & 278       & 91.50\%      & 0.89\%          & 93.20\%    & 2.59\%          \\
D2            & 90.60\%   & 616       & 91.32\%      & 0.72\%          & 95.02\%    & 4.42\%          \\
D3            & 89.73\%   & 628       & 90.26\%      & 0.53\%          & 94.51\%    & 4.78\%          \\
D4            & 91.70\%   & 915       & 92.44\%      & 0.74\%          & 95.59\%    & 3.89\%          \\
D5            & 86.11\%   & 34        & 89.72\%      & 3.61\%          & 94.44\%    & 8.33\%          \\
D6            & 52.65\%   & 94        & 58.34\%      & 5.69\%          & 63.01\%    & 10.36\%         \\
D7            & 86.84\%   & 994       & 87.96\%      & 1.12\%          & 91.67\%    & 4.83\%          \\
D8            & 69.95\%   & 1,273     & 72.34\%      & 2.39\%          & 76.98\%    & 7.03\%          \\ \midrule
\textbf{Avg.} & \textbf{} & \textbf{} & \textbf{}    & \textbf{1.96\%} & \textbf{}  & \textbf{5.78\%} \\ \bottomrule
\end{tabular}
\end{table}

\subsection{Effectiveness of Testability-Aware Attention Mechanism} \label{SEC:EXP:Attn}

To demonstrate that our testability-aware attention mechanism can capture more helpful information in the TPI task, we design an ablation study with different aggregation configurations. 
We apply three different aggregators in our RL agent: the aggregator without attention (w/o Att.), the aggregator with normal logic-aware attention, same as~\cite{li2021representation, abs-1904-11088} (Nor. Att.) and the aggregator with our proposed testability-aware attention as Sec.~\ref{Sec:Method:Attention} (DeepTPI Att.). Except for the above aggregation function, the remaining RL components keep the same hyperparameters.

\begin{table}[!t]
\renewcommand\tabcolsep{2.5pt}
\centering
\caption{The Test Coverage Improvement Comparison of Different Aggregator} \label{TAB:attn}
\begin{tabular}{@{}lc|cc|cc|cc@{}}
\toprule
              & TC               & w/o Att. & Imp.            & Nor. Att. & Imp.            & DeepTPI Att. & Imp.    \\ \midrule
D1            & 90.61\%          & 92.62\%       & 2.01\%          & 92.75\%      & 2.14\%          & 93.20\%    & 2.59\%  \\
D2            & 90.60\%          & 91.31\%       & 0.71\%          & 92.04\%      & 1.44\%          & 95.02\%    & 4.42\%  \\
D3            & 89.73\%          & 91.60\%       & 1.87\%          & 90.79\%      & 1.06\%          & 94.51\%    & 4.78\%  \\
D4            & 91.70\%          & 93.67\%       & 1.97\%          & 92.84\%      & 1.14\%          & 95.59\%    & 3.89\%  \\
D5            & 86.11\%          & 87.67\%       & 1.56\%          & 90.21\%      & 4.10\%          & 94.44\%    & 8.33\%  \\
D6            & 52.65\%          & 58.29\%       & 5.64\%          & 59.45\%      & 6.80\%          & 63.01\%    & 10.36\% \\
D7            & 86.84\%          & 89.84\%       & 3.00\%          & 90.76\%      & 3.92\%          & 91.67\%    & 4.83\%  \\
D8            & 69.95\%          & 75.68\%       & 5.73\%          & 75.87\%      & 5.92\%          & 76.98\%    & 7.03\%  \\ \midrule
\textbf{Avg.} & \textbf{82.27\%} & \textbf{}     & \textbf{2.81\%} & \textbf{}    & \textbf{3.32\%} & \textbf{}  & \textbf{5.78}\%  \\ \bottomrule
\end{tabular}
\end{table}

Tab.~\ref{TAB:attn} shows the experimental results. The test coverage improvement between post-TPI and the pre-TPI circuit is denoted as \text{Imp.}. 
The RL agent without any attention mechanism only achieves $2.81\%$ Imp. on average. This setting performs worst because it cannot determine the impact of detecting faults of different neighbors. 
Although normal attention can distinguish the dominant neighbors, it cannot capture the synergistic effect on testability between predecessors and successors. 
The normal attention also has an inferior performance to our proposed testability-aware attention mechanism. The former only improves $3.32\%$ test coverage on average, lower than $5.78\%$ by our proposed attention mechanism. 

To summarize, the testability-aware attention is more efficient on the TPI task because it not only captures the dominance of neighbors but also considers the coordination between the nodes benefiting fault sensitizing and the nodes in the easiest fault propagation path. As a result, the agent with testability-aware attention increases $2.06\times$ TC improvement than the agent without attention. 

\subsection{Effectiveness of Feature-Based Pre-Training Approach}\label{SEC:EXP:pretrain}

To investigate the effectiveness of the feature-based pre-training scheme, we train an RL agent without the pre-trained DeepGate model (abbreviated as \text{w/o Pre-Train}). Different from including the node embedding by the pre-trained DeepGate model, the initial node feature for \text{w/o Pre-Train} only contains the $3$-d one-hot encoding of gate types. 
Tab.~\ref{TAB:pretrain} shows the test coverage improvement (Imp.) by these two agents with the same number of inserted TPs.
As the customized agent (w/o Pre-Train) lacks the prior logic circuit knowledge, it only improves $2.93\%$ test coverage on average, which is much lower than $5.78\%$ Imp. obtained by the original agent (w/ Pre-train). 

To conclude, the pre-trained model provides a more helpful gate representation. Such abundant representation can help the agent achieve about $1.97\times$ test coverage improvement with the same TPI overhead.

\begin{table}[!t]
\renewcommand\tabcolsep{2.5pt}
\centering
\caption{The Test Coverage Improvement Comparison between w/o Pre-train and w/ Pre-train} \label{TAB:pretrain}
\begin{tabular}{@{}lcc|cc|cc@{}}
\toprule
              & TC        & \# TPs       & w/o Pre-Train & Imp.            & w/ Pre-train & Imp.            \\ \midrule
D1            & 90.61\%   & 278       & 91.63\%         & 1.02\%          & 93.20\%      & 2.59\%          \\
D2            & 90.60\%   & 616       & 92.26\%         & 1.66\%          & 95.02\%      & 4.42\%          \\
D3            & 89.73\%   & 628       & 91.42\%         & 1.69\%          & 94.51\%      & 4.78\%          \\
D4            & 91.70\%   & 915       & 93.18\%         & 1.48\%          & 95.59\%      & 3.89\%          \\
D5            & 86.11\%   & 34        & 89.23\%         & 3.12\%          & 94.44\%      & 8.33\%          \\
D6            & 52.65\%   & 94        & 57.49\%         & 4.84\%          & 63.01\%      & 10.36\%         \\
D7            & 86.84\%   & 994       & 90.52\%         & 3.68\%          & 91.67\%      & 4.83\%          \\
D8            & 69.95\%   & 1,273     & 75.88\%         & 5.93\%          & 76.98\%      & 7.03\%          \\ \midrule
\textbf{Avg.} & \textbf{} & \textbf{} & \textbf{}       & \textbf{2.93\%} & \textbf{}    & \textbf{5.78\%} \\ \bottomrule
\end{tabular}
\end{table}

\subsection{Analysis of Computational Complexity} \label{SEC:EXP:Comp}
The proposed DeepTPI utilizes modern GPUs for RL training and inference. To give an idea of the computational complexity, we take circuit D1 as an example. The DeepTPI takes $132.07$ seconds to insert 10 TPs on D1 (running on Nvidia GeForce GTX 1080 Ti GPU), while the commercial DFT tool takes $12.79$ seconds to insert the same number of TPs. Overall, DeepTPI takes around $10\times$ more runtime than the commercial tool. As TPI is a one-time effort process in the DFT phase, the computational overhead of DeepTPI is acceptable.

\begin{figure}[t!]
	\centering
	\includegraphics[width=0.4 \textwidth]{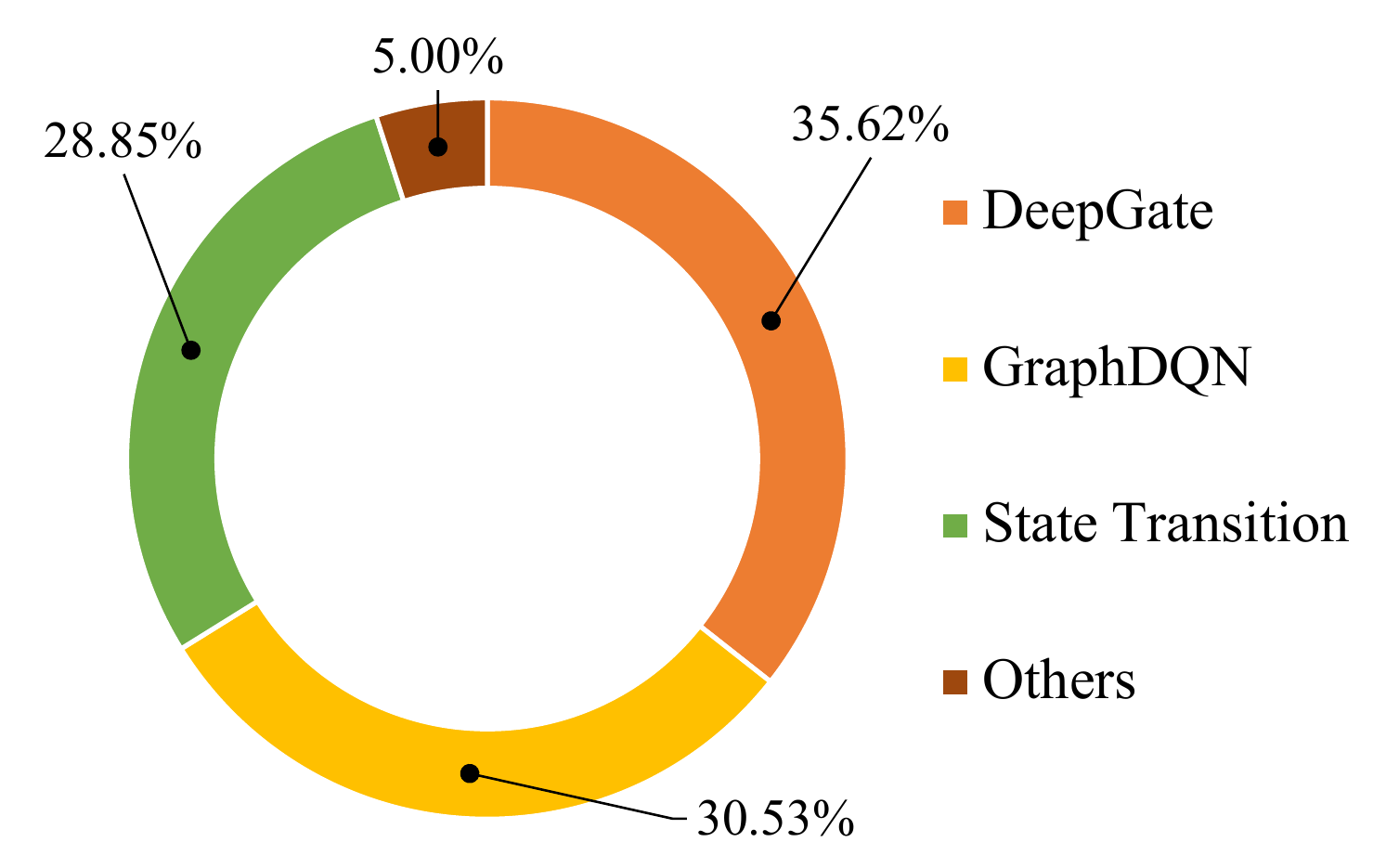}
	\caption{The Runtime Breakdown of the DeepTPI Inference}
	\label{FIG:Time}
\end{figure}

We also measure the runtime for different modules of DeepTPI. Fig.~\ref{FIG:Time} shows the runtime breakdown of the DeepTPI inference on the circuit D1. The pre-trained DeepGate and Graph-DQN take $35.62\%$ and $30.53\%$ out of the total time, respectively. 
Although the Graph-DQN is more complex than DeepGate (see Tab.~\ref{Tab:FLOPs}), DeepGate consumes more time as it updates the node hidden state with the orders of the logic level. 
Contrary to aggregating messages level-by-level, the Graph-DQN updates all nodes over the whole graph simultaneously.  
Another time-consuming computation process is the State Transition, which takes $28.8\%$ out of total time. The State Transition refers to the re-calculation of both the forward logic level (from PI to PO) and backward logic level (from PO to PI) after inserting one TP. It should be noted that maintaining logic level only profits the pre-trained model inference, as DeepGate aggregates the message level-by-level. 

There are several potential ways to further improve the efficiency of DeepTPI. For example, DeepTPI can be paralleled on multi-devices, which can reduce the runtime. The number of GNN iterations can also be reduced with more training data. Or we can insert multiple TPs in each inference round. We leave them for future work.

\section{Conclusion} \label{Sec:Conclusion}
This paper proposes DeepTPI, a test point insertion approach with deep reinforcement learning.
As the circuit can be naturally modeled as a graph and the action is defined as inserting TP on the discrete node over the graph, we train an agent with the combination of deep Q-learning and a graph neural network (Graph-DQN) to approximate the value of all possible actions. We also embed a testability-aware attention mechanism into the Graph-DQN to capture the dominant impact on testability from both predecessors and successors. The optimal policy is to conduct the action with the maximum predicted value. 
Besides, we apply DeepGate as a pre-trained model to learn the universal gate embedding as the prior circuit information. The experimental results prove the effectiveness of our strategies and demonstrate that DeepTPI can achieve better performance than the commercial DFT tool.

\balance
\bibliographystyle{IEEEtran}
\bibliography{ref}

\end{document}